\documentclass[letterpaper, 10 pt, conference]{ieeeconf}  
                                                          
\usepackage[utf8]{inputenc}
\usepackage[english]{babel}
\usepackage{amsmath,empheq,amssymb}
\usepackage{mdframed}
\newmdtheoremenv{problem}{Problem}
\newcommand{\norm}[1]{\left\lVert#1\right\rVert}

\usepackage{graphicx}
\usepackage[version=4]{mhchem}
\usepackage{siunitx}
\usepackage{longtable,tabularx}
\usepackage[backend=bibtex,style=numeric,sorting=none]{biblatex}
\title{Minimum time trajectory planning for surveying using UAVs}
\author{Srinath Tankasala, Can Pehlivanturk and Mitch Pryor}

\begin{document}

\maketitle
\begin{abstract}
    In this paper, we present a motion planning strategy for UAVs that generates a time-optimal trajectory to survey a given target area. There are several situations where completing an aerial survey is time sensitive, such as gaining situational awareness for first responders, surveying hazardous environments, etc. One of the challenges in such cases is to plan a time-optimal trajectory for the drone. To this end, we present an autonomous aerial survey framework that minimizes the time taken to completely explore a target area or volume using drones. In this work, (i) we present an approach, where for a known flight survey pattern, the planner can generate time-optimal flight paths in 3-D (ii) we frame the planning problem as a discrete non-linear program, and reduce the time taken to compute its solution by using an SOCP relaxation (iii) The given path is then executed using a simple trajectory tracking controller on a quadrotor to demonstrate its capability on hardware.
\end{abstract}
\section{Introduction}

Autonomous Unmanned Aerial Vehicles (UAVs) are powerful tools with valuable applications across many industries and disciplines. They are being used for delivery, inspections, racing and cinematography, etc. In recent years, industrial inspection has undergone significant automation with advancements in mobile robotics, such as the Boston Dynamics Spot robot. For outdoor inspections and surveys, UAVs are still the most popular choice as they are unrestricted in 3D space and more capable to reach hard to access areas and remote locations. Aerial surveying and scene reconstruction is an application where the UAVs are tasked with capturing visual or sensory information about a Region of Interest (RoI). Aerial surveying requires the generation of a flight path/plan that covers the given RoI. This paper presents an autonomous aerial survey framework to generate time-optimal trajectories to cover a target area using drones while imposing constraints to ensure good quality of data collection.  

\begin{figure}[h]
    \centering
    \includegraphics[width=3in]{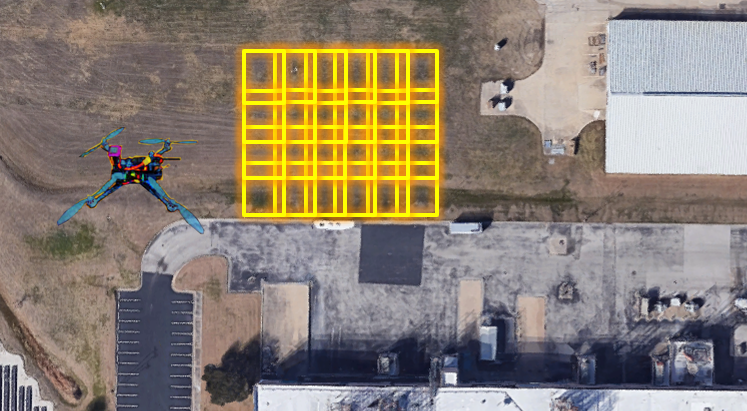}
    \caption{Representative pic of the drone and discretization of of survey area with defined image overlap (yellow rectangles) for stitching}
    \label{fig:overlapsurvey}
\end{figure}

Both Coverage Path Planning (CPP) \cite{CPPSurvey} and time-optimal trajectory generation \cite{MPC:Eren, traj:gusto, traj:mao} are well studied problems in the literature. The aerial surveying problem uses principles from CPP \cite{choset1998coverage,acar2006sensor} to generate a set of desired waypoints that need to be visited by the agent and this is usually implemented as a global planner, see Figure \ref{fig:overlapsurvey}. Minimum-time trajectory generators are then used to plan the path passing though the waypoints obtained from the global planner. Recent works have made great progress in minimum-time trajectory generation for navigating across a known set of waypoints \cite{lai2006time, foehn2021time, spedicato2017minimum, traj:benchmark}. However, they are not necessarily time-optimal for collecting visual survey data. When doing a visual/thermal imaging survey, the agent must pass through a given set of viewpoints (from a global planner) below a certain speed threshold to avoid motion blur in the captured images. For surveying tasks this motion-blur constraint need only be applied when the UAV passes through a desired waypoint, instead of the entire trajectory. The speed threshold can be determined based on the quality of reconstruction needed, i.e. high motion blur would lead to low quality reconstruction.

In this paper, we develop a minimum-time trajectory generation framework that is suitable for surveying using drones. We approximate the dynamics of the drone to a point mass in order to frame the problem as a Non-Linear Program (NLP) that can be solved with standard optimization solvers, ex. IPOPT \cite{IPOPT}. The essence of a UAV's dynamics can be captured by a point mass approximation within a certain velocity bound ($\bar{v}$) and can be used for motion planning, \cite{foehn2021alphapilot} is one notable example where this approximation was used to achieve good results. The velocity bound ($\bar{v}$) is usually higher than the allowable speed for motion blur ($v_{blur}$) when surveying (the calculation of $v_{blur}$ is described in \ref{sec:theory}). We then demonstrate the feasibility of our trajectory generation method by completing a survey of a target RoI on a custom-built drone. 

We structure the rest of the paper as follows. In Section \ref{sec:related work} we give an overview of different minimum-time trajectory generation methods in the literature. In Section \ref{sec:theory} we provide the formulation of the surveying problem and the methods employed for solving it. Section \ref{sec:simulations} shows simulation results of the trajectory generator along with comparisons to other trajectory generation algorithms in the literature. In Section \ref{sec:exp} we present the experimental flight tests from implementing the algorithm on a custom-built drone. Finally, we provide our conclusions and future work and possible improvements in Section \ref{sec:conclusions}.

\section{Related Work}
\label{sec:related work}
Numerous view path planning algorithms algorithms have been developed in the literature for scene reconstruction. They are largely focused on efficient viewpoint selection to achieve maximum coverage of the scene. Notable examples include \cite{choset2001coverage, andersen2014path}. These coverage planners compute camera viewpoints of the RoI and the pattern in which they will be visited (zig-zag, spiral, etc.). The computed viewpoint tour is then fed to a local planner to generate a continuous trajectory. Formulating a minimum-time local planner for surveying is the main focus of this paper and there are several approaches to achieve this in the literature. In this work, a quadrotor is used as the surveying agent.

The differential flatness property of quadrotors \cite{mellinger2011minimum} makes it sufficient to only plan the position, velocity, acceleration and yaw of the drone. The rotational dynamics like pitch and roll rate, are accounted for by applying constraints on higher order derivatives like jerk and snap \cite{mellinger2011minimum} etc. Considering only the translational dynamics, a quadrotor is a point mass with an upper bound on the magnitude of thrust making the input space a hypersphere. When the input space is a hypersphere there always exists a time-optimal control input that attains its values in the boundary of the input value set. This is known as a boundary control strategy, \cite{du2018boundary}, or bang-bang control if the input set is a hypercube. There exist several approaches in the literature that use similar strategies for generating time-optimal trajectories for UAVs: Lai et al \cite{lai2006time} uses a discretizes in the time domain and input function to formulate the quadrotor steering problem as an non-linear program (NLP) that can be solved with the genetic algorithm. In Hehn et al \cite{traj:benchmark} the authors simplify the quadrotor dynamics to a 2D model and apply Pontryagin's Minimum Principle (PMP) to solve for the minimum-time solution. 

The bang-bang strategy is a popular choice for minimum-time trajectories \cite{lupashin2011adaptive,traj:benchmark} and it provides a solution where the system uses maximum actuator input along the boundary of the admissible control envelope (hypercube). Quadrotors are however underactuated systems and hence not all of the available thrust can used for accelerating the vehicle. Some actuator input is used to align the vehicle thrust in the correct direction. Nevertheless, the point mass approximation can be considered valid within a certain velocity bound. The point mass minimum-time solution can also serve as an initial/low fidelity solution in many planning strategies \cite{foehn2021time, ryou2020multi}. To this end, we demonstrate our proposed trajectory generation strategy by applying it to solve a minimum-time surveying task. The trajectory generation method in this work can be applied to other tasks as well such as drone racing, etc. 

\section{Surveying methodology}
\label{sec:theory}

To understand the requirements of an inspection survey we discuss the problem setup and assumptions. Typical industrial survey problems are offline path planning problems where information about the RoI is known \textit{a priori}. Hence we can generate viewpoints for full coverage of the RoI using cellular decomposition. A cell can be defined as the area or volume that will be explored in a single measurement, and the size of the cell is determined by the sensing requirements. For example, in a photogrammetry application, a cell can be defined as the footprint of the camera, and the size and shape of the cell can depend on factors such as the aspect ratio of the camera, the required ground resolution, and the required overlap between the photographs. The target area is then decomposed into cells of predefined size \cite{cvrg:uavsurvey}. To determine the flight survey parameters, we first calculate the flight height ($H$) based on the desired Ground Sample Distance (GSD) using the formula in \cite{kakaes2015drones}. Next we calculate the maximum allowable speed of the drone ($v_{blur}$) while passing a waypoint. This can be computed by the formula in equation \eqref{eq:vblur_formula}.

\begin{align}
    v_{blur} = \frac{\rho}{G_x T_s} \label{eq:vblur_formula}
\end{align}
where $\rho$ (pixels) is the maximum allowable pixel blur, $G_x$ (pixels/m) is the ground resolution and $T_s$(sec) is the camera shutter speed. Allowable pixel blur $\rho$ can be determined by inspecting the reconstruction of a given scene. If the reconstruction needs to be improved, then a lower blur ($\rho$) should be selected by the user for surveying.

\subsection{Minimum time trajectory formulation}
In this work, we assume the order in which the waypoints are visited is known or given by the user, such as a zig-zag pattern \cite{cvrg:roboticssurvey} or from solving the Traveling Salesman Problem (TSP). The local planner must ensure the generated continuous trajectory (position, velocity, acceleration and yaw) passes through the desired waypoints in the least possible time while avoiding motion blur. For simplification purposes, the drone's yaw is assumed to be a constant (zero) at all points in the trajectory. This is reasonable as any choice of yaw orientation, $\psi$, of the drone does not distort pictures taken of the ground. In most image capture applications, being stationary at the capture point may not be necessary, so having an upper bound on the speed is sufficient to stay below a desired maximum pixel blur. We model the UAV dynamics as a double integrator. For such a system, time-optimal trajectory generation problem is of the form:

\begin{problem}\label{prb:general} 
Time optimal coverage trajectory generation
\[
\begin{gathered}
  \begin{aligned}
  &\min \int_{0}^{t_N} dt \\ 
  \mathrm{s.t.:}\ \ &\dot{{r}}(t) = {v(t)}, \quad {\dot{v}(t)} = {u(t) + g}, \\
  &r(t_i) = w_i ,\\
  \quad &\norm{v(t_i)} \leq v_{blur}, \forall\ i  = 1,...,N\\
  & v_{min} \leq  v(t) \leq v_{max}, \forall\ t
  \end{aligned}
\end{gathered}
\]
\end{problem}

where ${r(t)}\in \mathcal{R}^3$ and ${v(t)}\in \mathcal{R}^3$ are the position and velocity of the drone respectively at time ${t}$, ${u(t)}$ is the control input at time ${t}$ and $\bar{{u}}$ and ${g}$ are non-negative constants.  The constraints $r(t_i) = w_i$ and $\norm{v(t_i)} \leq v_{blur}$, correspond to the imposed waypoint position and speed constraints respectively. Additionally we impose velocity constraints on $v(t)$ along each individual axis, i.e. $v_{min}, v_{max} \in R$,  to satisfy the point mass assumption, similar to \cite{foehn2021alphapilot}. 

\textit{It should be noted that if $v_{blur}>\sqrt{3}v_{max}$, then the motion blur constraint, $\norm{v(t_i)} \leq v_{blur}$, becomes redundant. However we show later in section \ref{sec:simulations} that even in this case, our formulation performs better than other methods in the literature, like bang-bang for example.}

\subsection{Discretized time-optimal trajectory problem}

We solve the minimum-time solution of problem \ref{prb:general} by discretizing the time domain and framing it as an optimal control problem. The first step is to represent the continuous input function $u(t)$. The input function can be approximated by using a finite dimensional discretization. Examples include piecewise linear or piecewise polynomial of order $n$ representations \cite{traj:minsnap}. 

We know from Pontryagin's minimum-time principle \cite{athans2013optimal} that the optimal input lies on the boundary of a hypersphere for the point mass steering problem. Hence we assume input primitives that have magnitudes equal to the maximum allowable thrust ($\bar{u}$). In the discretized domain, the input vector switches from one value to the next at the switching time. If we use only one intermediate switching point, then the input between two waypoints is of the form:

\begin{equation}\label{eq:constinput}
 u(t) = 
  \begin{cases} 
   \bar{u} \frac{\eta_{n,1}}{\norm{\eta_{n,1}}} &  t_n \leq t \leq t_{n,s} \\
   \bar{u} \frac{\eta_{n,2}}{\norm{\eta_{n,2}}} &  t_{n,s} < t < t_{n+1}
  \end{cases}
\end{equation}

where $n \in \{1,2,...,N\}$ is the waypoint index, $t_n$ denotes the time at the $n^{th}$ waypoint and $t_{n,s}$ is the switching time where the input switches from one constant vector to another. The control input can thus be represented in a discrete form as follows:

\begin{equation}\label{eq:discreteinput}
 u_k = \left[\bar{u} \frac{\eta_{1,1}}{\norm{\eta_{1,1}}} ,  \bar{u} \frac{\eta_{1,2}}{\norm{\eta_{1,2}}}, ... , \bar{u} \frac{\eta_{N-1,1}}{\norm{\eta_{N-1,1}}}, \bar{u} \frac{\eta_{N-1,2}}{\norm{\eta_{N-1,2}}}\right] 
\end{equation}\\

with \textit{variable} discrete time steps:

\begin{equation}\label{eq:discretetimestep}
 dt_k =\left[t_{1,s}-t_1,t_{2}-t_{1,s},...,t_{N-1,s}-t_{N-1}, t_N-t_{N-1,s} \right]
\end{equation}

where $k \in \{1,2,...,2(N-1)\}$. The velocities and positions at their respective time steps can be determined through integration:

\begin{equation}\label{eq:discreteintegral}
\begin{aligned}
    r_{k+1} &= r_k + v_k dt_k + u_k\frac{dt_k^2}{2}\\
    v_{k+1} &= v_k + u_k dt_k
\end{aligned}
\end{equation}

The above constraints can then be consolidated into a non-linear program similar to Lai et al \cite{lai2006time}. The consolidated NLP in problem \ref{prb:nlp} is then solved to find the required waypoint velocities.

\begin{problem}\label{prb:nlp}
Discrete time-optimal coverage NLP
\[
\begin{gathered}
    min \sum_{k = 1}^{2(N -1)} dt_k\\
  \begin{aligned}
  &r_{k+1} = r_k + v_k dt_k + u_k\frac{dt_k^2}{2}\\
  &v_{k+1} = v_k + u_k dt_k\\
  &r_j = wp_n ,\quad  n  = 1,...,N, \\
  &\norm{v_i} \leq v_{blur}, \quad \forall\ i = 1, 3, ..., 2N-1\\
  &\norm{u_k} = \bar{u} \\
  &-\bar{v} \leq v_j \leq \bar{v} \quad \forall\ j=1,2,3,...,2N-1
  \end{aligned}
\end{gathered}
\]
\end{problem}
where, $v_j\in \mathbb{R}^3$ and $-\bar{v} \leq v_j \leq \bar{v}$ is computed for each axis individually. If required, we can impose a lower bound on $u_z$ (global Z-direction) to ensure a minimum thrust is always exerted thus maintaining non-zero rotor speeds.

Solving the NLP in problem \ref{prb:nlp} is computationally intensive due to the input magnitude equality constraint, $\norm{u_k} = \bar{u}$. The input magnitude is equality constrained to $\bar{u}$ to conform to Pontryagin's minimum principle (PMP). However, that can lead to large velocities if the distances between waypoints are large. For example, a 1D case with only 2 waypoints separated by a large distance would lead to a large switching times and high velocities. To avoid this, the NLP is first solved with the input magnitude as an equality constraint. If the solver fails to find a solution satisfying the bounds on velocities $\bar{v}$, then the input constraint is relaxed to an inequality constraint, i.e. $\norm{u_k} \leq \bar{u}$ and additional switching points are added.

To provide a starting point to the solver, Problem \ref{prb:nlp} is relaxed into a Second Order Cone Program (SOCP) by replacing the variable time steps $dt_k$ with a constant time step $dt$, relaxing the input constraint $\norm{u_k} \leq \bar{u}$, and defining a slack variable such that $\bar{u} \leq \sigma_k$.

\begin{problem}\label{prb:socp} 
SOCP relaxation of NLP
\[
\begin{gathered}
    min \sum_{k = 1}^{2(N -1)} \sigma_k\\
  \begin{aligned}
  &r_{k+1} = r_k +  v_k dt + u_k \frac{1}{2}dt^2\\
  &v_{k+1} = v_k +  u_k dt\\
  &r_j = wp_n ,\quad \norm{v_i} \leq \bar{v}, \quad n  = 1,...,N\\
  &\norm{u_k} \leq \bar{u}, \quad i = 1, 3, ..., 2N-1 \\
  &\bar{u} \leq \sigma_k\\
  &-\bar{v} \leq v_j \leq \bar{v} \quad \forall\ j=1,2,3,...,2N-1
  \end{aligned}
\end{gathered}
\]
\end{problem}
where $-\bar{v} \leq v_j \leq \bar{v}$ is computed for each axis individually. The optimal $dt^*$ and velocities to the SOCP in problem \ref{prb:socp} is used as the initial guess to the NLP optimization, i.e. $[v_1^*, v_2^*, v_3^*, ..., v_{2(N-1)}^*]_{SOCP}$ is the initial guess for solving problem \ref{prb:nlp}.

The continuous time trajectory passing through the prescribed waypoint positions and the corresponding optimal waypoint velocities can then be determined by integrating the solution to Problem \ref{prb:nlp} between the time steps, as in equations \eqref{eq:interpolate}:

\begin{equation}\label{eq:interpolate}
\begin{aligned}
 u(t) &= u_k  \\
 v(t) &=  v_k +  u_k t\\
 r(t) &= r_k + v_k t +  \frac{u_k t^2}{2}\\
 k &=
 \begin{cases}
 1 & 0 \leq t \leq \sum_{k=1}^{1} dt_k\\
 2 & \sum_{k=1}^{1} dt_k \leq t \leq \sum_{k=1}^{2} dt_k\\
 \vdots & \vdots\\
 2N-1 & \sum_{k=1}^{2N-2} dt_k\leq t \leq \sum_{k=1}^{2N-1} dt_k
 \end{cases}
\end{aligned}
\end{equation}\\

\section{Simulation Results}
\label{sec:simulations}

The SOCP of problem \ref{prb:socp} is solved using ECOS \cite{ECOS}. A line search is performed on $dt$ to find a feasible solution that has the minimum final time possible with the constant $dt$ assumption. The solution for the SOCP is then used as the initial guess for the NLP. The NLP in problem \ref{prb:nlp} is solved using a linear solver (ma27) from the library IPOPT (Interior Point Algorithm for Nonlinear Optimization) \cite{IPOPT}.

\begin{figure}[!h]
    \centering
    \includegraphics[width=3.3in]{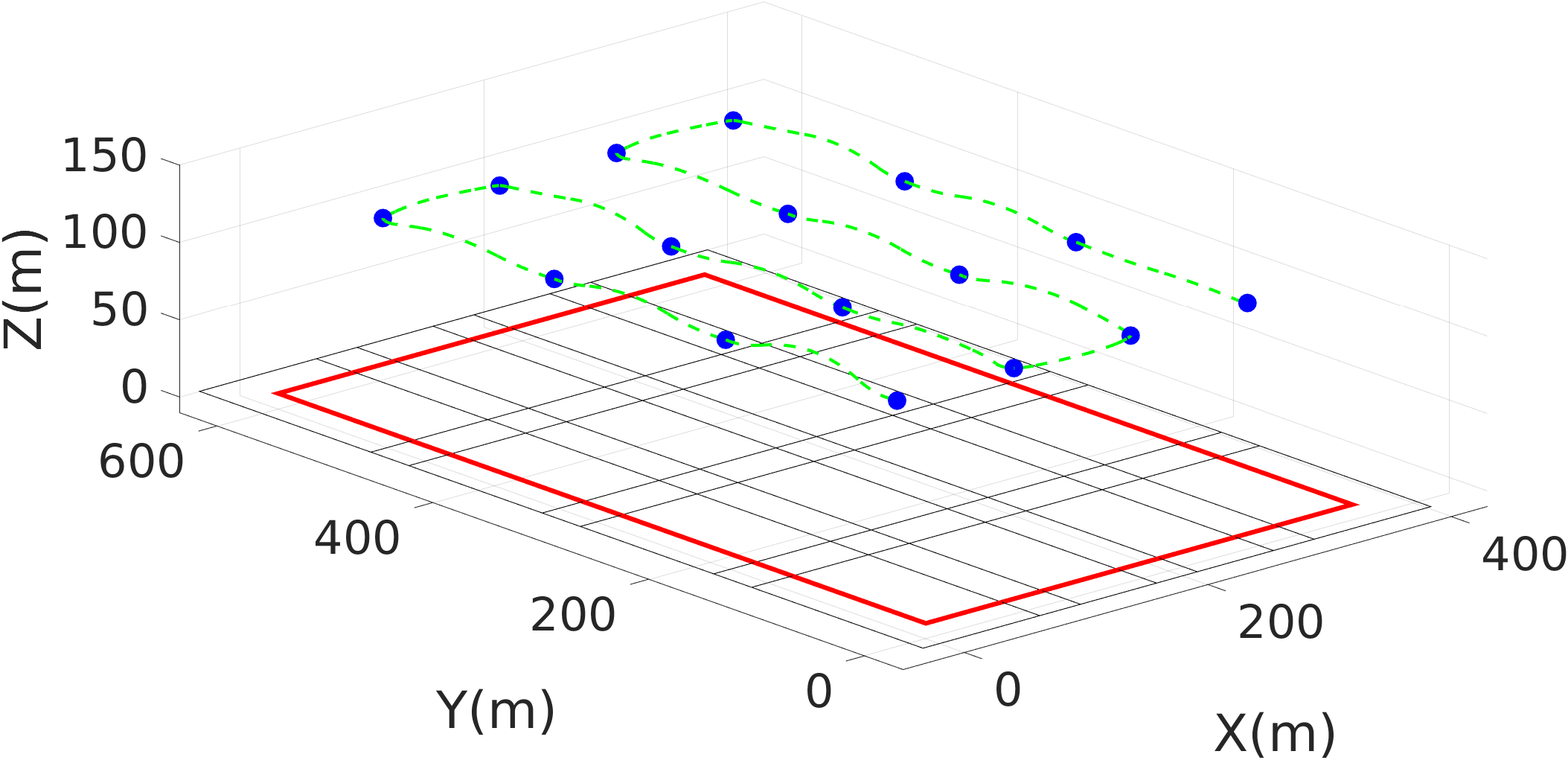}
    \caption{Generated trajectory (green) for RoI(red) with target waypoints (blue) and $v_{blur}=5m/s$}
    \label{fig:traj_3d}
\end{figure}

\begin{figure}[!h]
    \centering
    \includegraphics[width=3in]{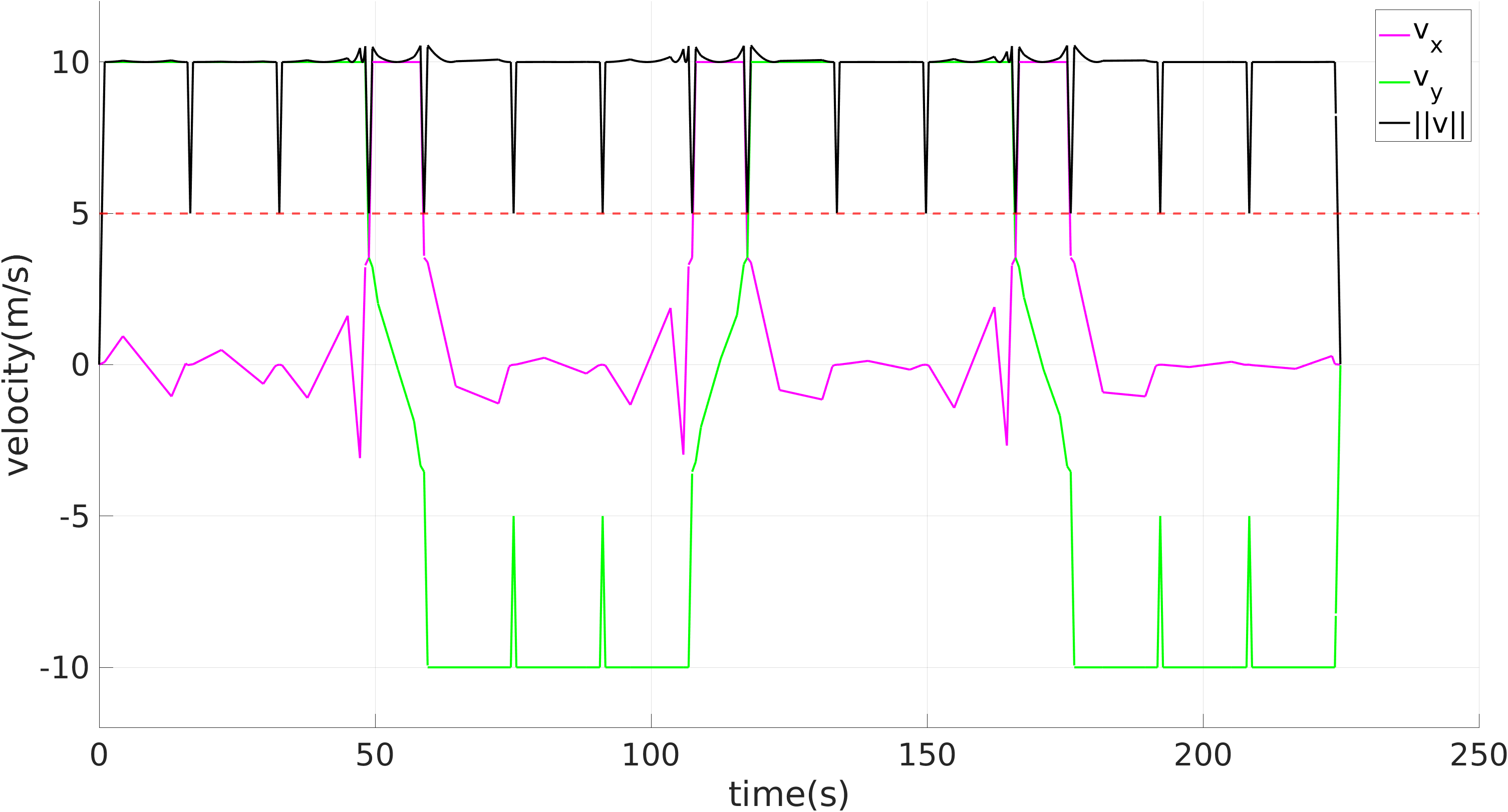}
    \caption{Velocity of the generated trajectory with $v_{blur}$ as $5m/s$}
    \label{fig:v_intermediate}
\end{figure}

The choice of $v_{blur}$, $\bar{v}$ and $\bar{u}$ depends on the quadrotors system parameters. For our simulation results we choose a specific thrust $\bar{u}=14 N/kg, \bar{v}=10m/s$ and $v_{blur}=5m/s$. The planner surveys an ROI of dimensions $350m\times 600m$ at an altitude of 120m. The trajectory generated from solving the NLP with the above parameters is shown in figure \ref{fig:traj_3d} and the corresponding velocities are shown in figure \ref{fig:v_intermediate}.

In figure \ref{fig:v_intermediate} it is clear that the trajectory satisfies the $v_{blur}$ speed constraint when passing through each waypoint. Also the velocity along each axis is bounded by $\bar{v}$.The sharp changes in velocity near the waypoints are difficult to track on a real drone and the trajectory may need to be smoothed before deploying to a real quadrotor. One possible smoothing technique is discussed later in Section \ref{smoothing}.

\subsection{Comparison with alternative trajectory generation methods}

In order to minimize computation time, most conventional UAV survey planners like Mission planner and Qgroundcontrol generate trajectories with linear velocity references along each grid line of the survey pattern. To alleviate the problem that the drone may not track these velocities, they add an additional turnaround distance at the end of each grid line, increasing the flight time and distance travelled by the drone. Since such surveying methods don't guarantee minimum flight time, we instead use the bang-bang solution in the literature \cite{lavalle2006planning} as a benchmark for evaluating our algorithm. Due to the upper bounds constraint on our generated velocities, we use bang-singular-bang inputs to represent the minimum-time solution. Clearly, the survey time increases as the value of $v_{blur}$ reduces (agent needs to slow down more), so we compare the best possible survey times using both methods, i.e. $v_{blur}=\infty$. We generate waypoints spanning $30m\times20m$ with a maximum specific thrust $\bar{u}=12N/kg$. We use 3 intermediate switching points for both methods (NLP and bang-bang).

\begin{figure}[ht]
    \centering
    \includegraphics[width=3in]{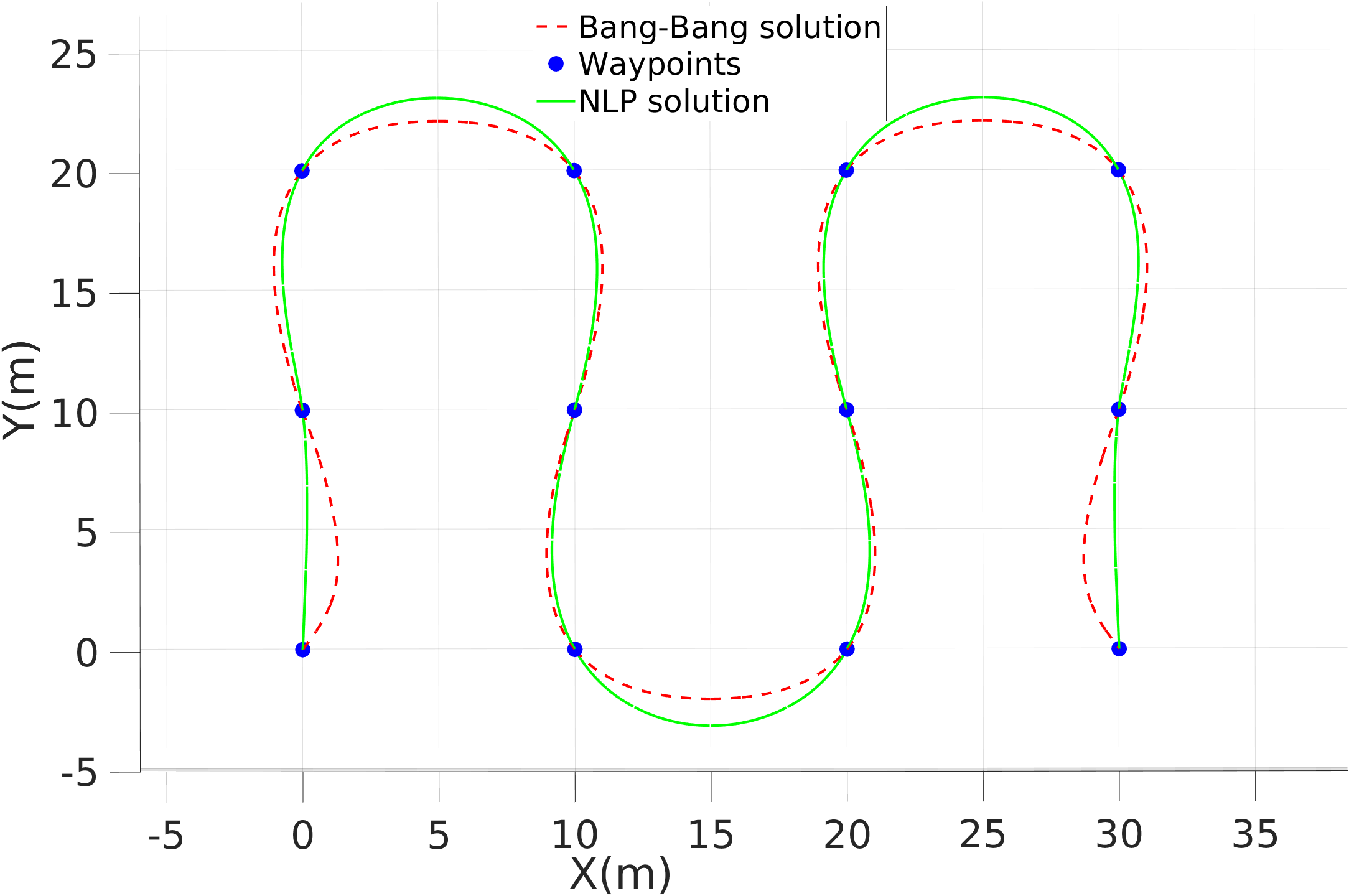}
    \caption{Comparison of bang-bang trajectory (solid) vs our trajectory generator (dotted)}
    \label{fig:traj_min_bang_comp}
\end{figure}

\begin{figure}[ht]
    \centering
    \includegraphics[width=3.14in]{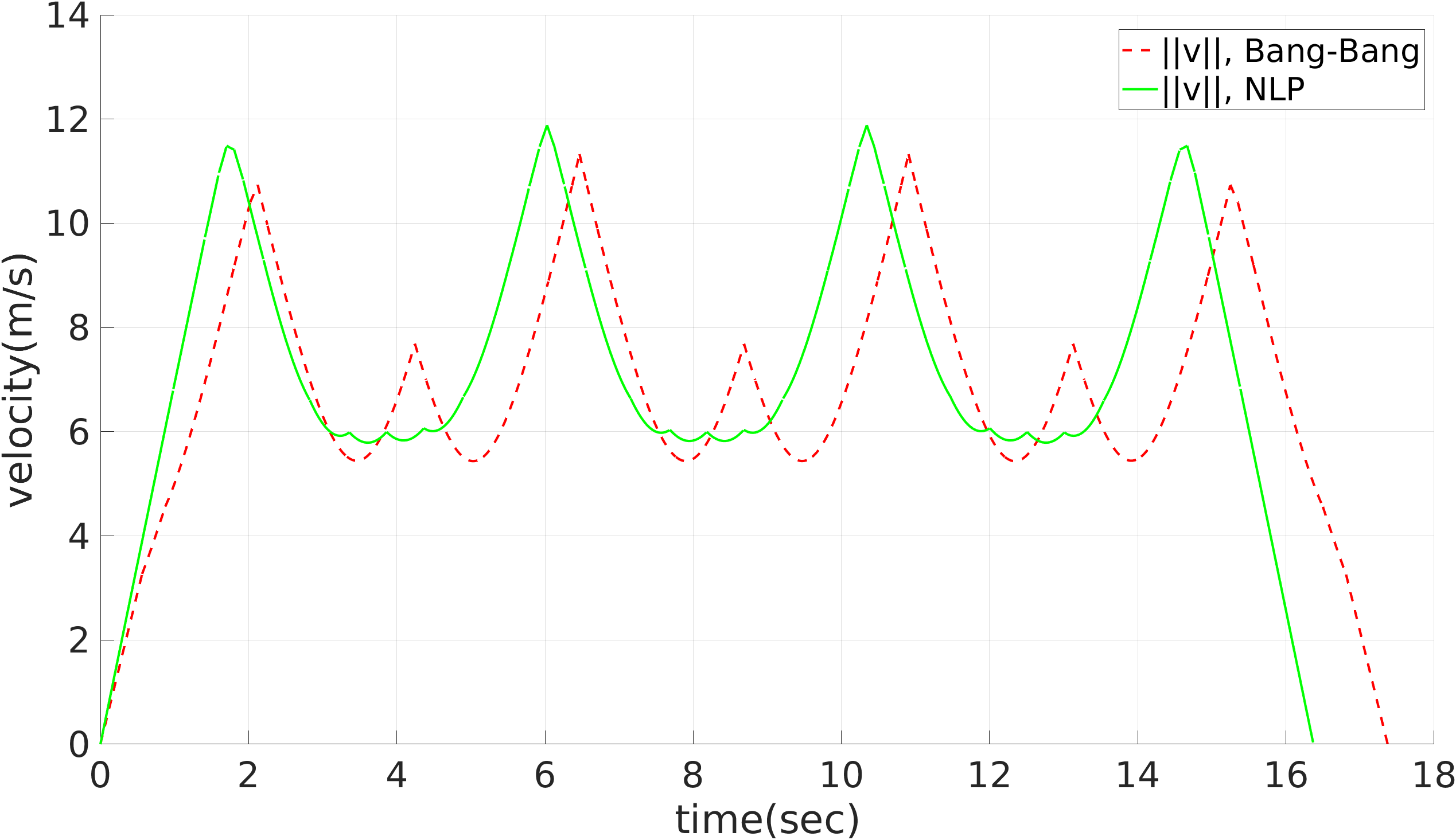}
    \caption{Comparison of bang-bang trajectory (solid) vs the NLP trajectory generator (dotted)}
    \label{fig:vel_min_bang_comp}
\end{figure}

\begin{figure}[ht]
    \centering
    \includegraphics[width=3.14in]{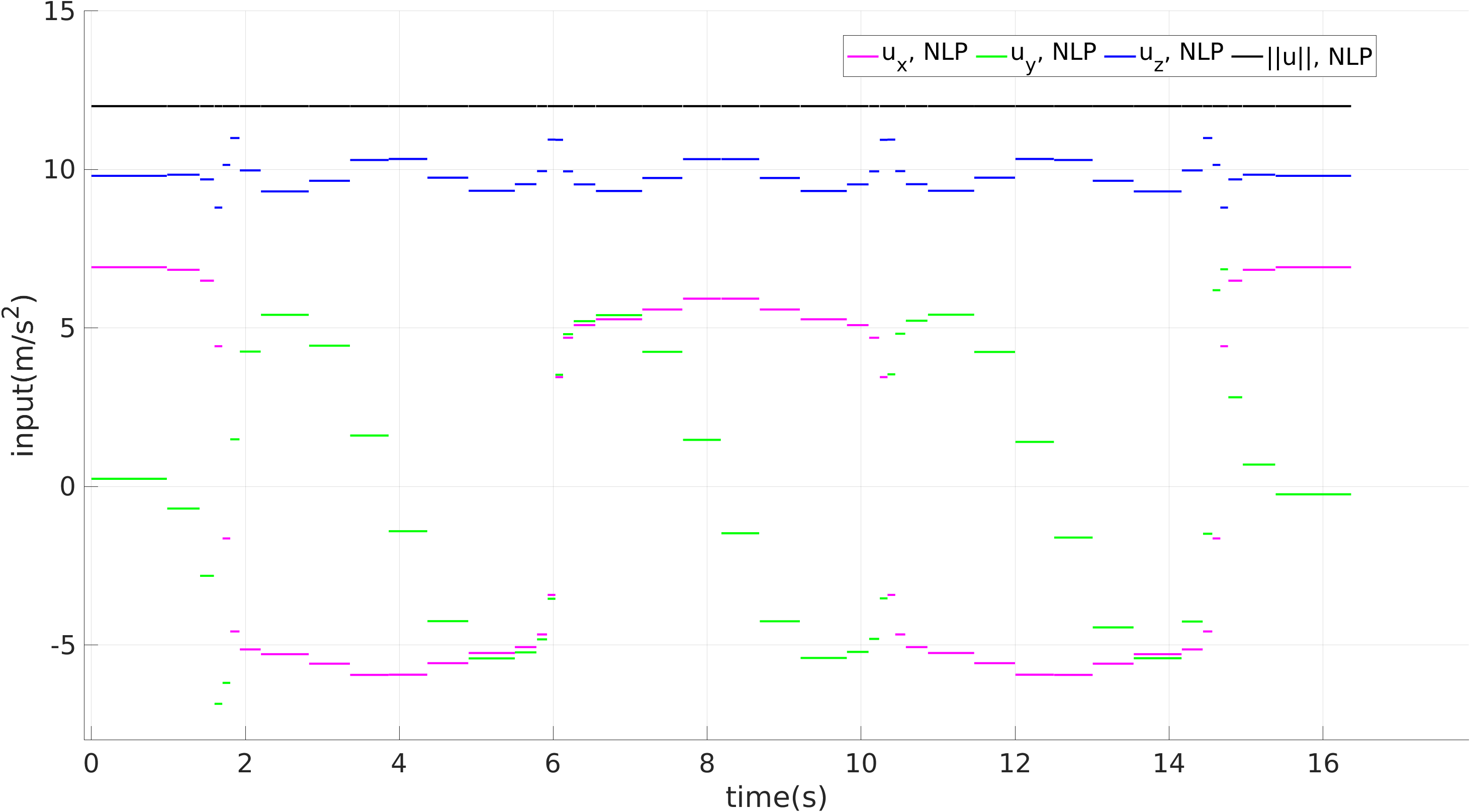}
    \caption{Time-optimal inputs from NLP trajectory generator (dotted)}
    \label{fig:inp_min_bang_comp}
\end{figure}

As can be seen in figures \ref{fig:traj_min_bang_comp} and \ref{fig:vel_min_bang_comp} our algorithm is able to complete the survey in 16.3s and the bang-bang solution takes 17.5s, which is $\approx8\%$ slower. Our algorithm performed faster for all other $v_{blur}$ values that were evaluated. The reason for better performance is that the input space for the NLP solution is the entire surface of the hypersphere whereas in the bang-singular-bang case it is restricted to the extremities of the input hypercube.
Figure \ref{fig:inp_min_bang_comp} shows the inputs obtained from solving the NLP. We can see that the inputs take values along the hypersphere, instead of switching between two values. This makes it explore a larger input space while ensuring that $\norm{u}$ is always at the upper bound, satisfying PMP. While $8\%$ doesn't save much time in this case, it becomes significant for larger survey areas with more waypoints and also in other applications such as drone racing. In terms of computation time, solving the NLP is made easier by using the SOCP approximation, making it comparably fast to the bang-bang formulation.

\subsection{Trajectory smoothing}
\label{smoothing}
Since the trajectory based on the point mass assumption may not be trackable for a quadrotor's dynamics, like in fig \ref{fig:v_intermediate}, the trajectory can be smoothed to make it feasible. We smooth our output by fitting 4th order polynomials between consecutive waypoints. The initial and final positions and velocities are used as references to fit the 4th order polynomial between consecutive waypoints. 

Thus given a minimum-time solution $r(t)$ from solving problem \ref{prb:nlp}, the smoothed trajectory between consecutive waypoints, $[k,\ k+1]$, $\bar{r}_k(t)$ can be written as:
\begin{align}
\bar{r}_k(T) &= a_{0,k} + a_{1,k}T + a_{2,k}T^2 + a_{3,k}T^3 + a_{4,k}T^4\\
\dot{\bar{r}}_k(T) &= a_{1,k} + 2a_{2,k}T + 3a_{3,k}T^2 + 4a_{4,k}T^3 \\
\ddot{\bar{r}}_k(T) &= 2a_{2,k} + 6a_{3,k}T + 12a_{4,k}T^2 
\end{align}

where, $k=1,2,3...,N-1$ and $T = t-t_k$ is the relative time in the $k$-th segment of the trajectory. We enforce the position and velocities to match the trajectory generated from solving the NLP. Additionally we impose that the polynomial should pass the position midway between consecutive waypoints, i.e.:

\begin{equation}
    \bar{r}_k\left(\frac{dt_k}{2}\right) = r\left(\frac{t_k+t_{k+1}}{2}\right)
\end{equation}

The resulting smooth polynomial trajectory can then be deployed onto a drone. It should be noted that this smoothed trajectory doesn't necessarily ensure that the input magnitude remains $\bar{u}$ at all times. To ensure that the smoothened $u(t)$ has magnitude $\bar{u}$ at all times, please refer to our work in \cite{srinath2022smooth}.

\section{Experimental evaluation}
\label{sec:exp}

To demonstrate the algorithm's real world surveying capability, we utilized a quadrotor built with a PX4 autopilot and an NVIDIA Jetson Tx2 as the onboard computer. The quadrotor had a maximum specific thrust $\bar{u} = 14.4N/kg$. The quadrotor is tasked with surveying a simple rectangular area using a GoPro Hero 8 camera. 

The relevant camera parameters for the survey are below:

\begin{table}[!h]
\centering
\begin{tabular}{ |p{3cm}|p{3cm}|  }
\hline
 Variable name& Value\\
 \hline
 Total area & 40x30 = 1200 sq.m   \\
 Horizontal overlap&   50\% \\
 Horizontal FoV&   $87^0$ \\
 Vertical FoV&    $71^0$\\
 Total flight time & 23s \\
  Flight altitude &   10m \\
  Focal length &   20mm\\
 \hline
 \end{tabular}
\caption{List of experimental parameters}
\label{tab:flight_param}
\end{table}
Using the parameters in table \ref{tab:flight_param}, we generated the minimum-time trajectory from solving the NLP formulation. The controller implementation in \cite{marcelino_github} based on \cite{lee2010geometric} was used for trajectory tracking control in this work. This controller doesn't account for wind disturbances as that is not the primary focus of the paper. To take wind disturbance into account a suitable control method such as \cite{bisheban2018geometric, wang2020controller} can be used. The PID controller of \cite{lee2010geometric} is chosen as it is easier to tune and has fewer parameters compared to controllers that account for wind disturbances.

To make sure that the drone can follow the trajectory generated by the planner, we generate our trajectories \textbf{using $\bar{u} = 10.5N$}. The remaining specific thrust ($\approx 4N/kg$) is used for overcoming any disturbances. After tuning the flight controller, a maximum velocity along any axis $\bar{v}$ of $12m/s$ was found to be satisfactory.

\begin{figure}[!h]
    \centering
    \includegraphics[width=1.2in]{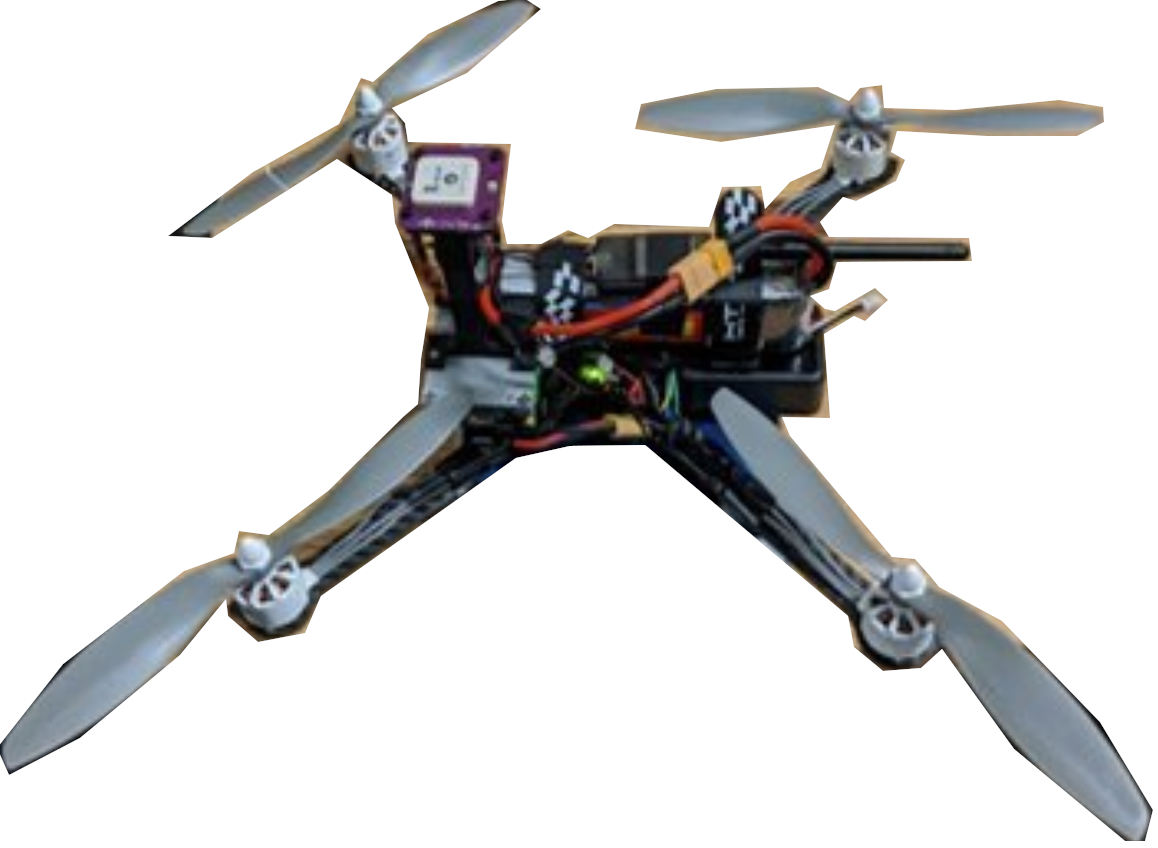}
    \caption{Custom-built drone with NVIDIA Jetson Tx2 and PX4 flight stack used for testing}
    \label{fig:drone_pic}
\end{figure}

The custom built drone, shown in fig \ref{fig:drone_pic}, had a mass of 1.15 kg. The tuned controller tracks the reference minimum-time trajectories based on our formulation. Based on the camera and reconstruction software (openDroneMap) used, $v_{blur}=8m/s$ was found to work best. It can be changed to accommodate any camera setting and model.
\\

\begin{figure}[h]
    \includegraphics[width=3in]{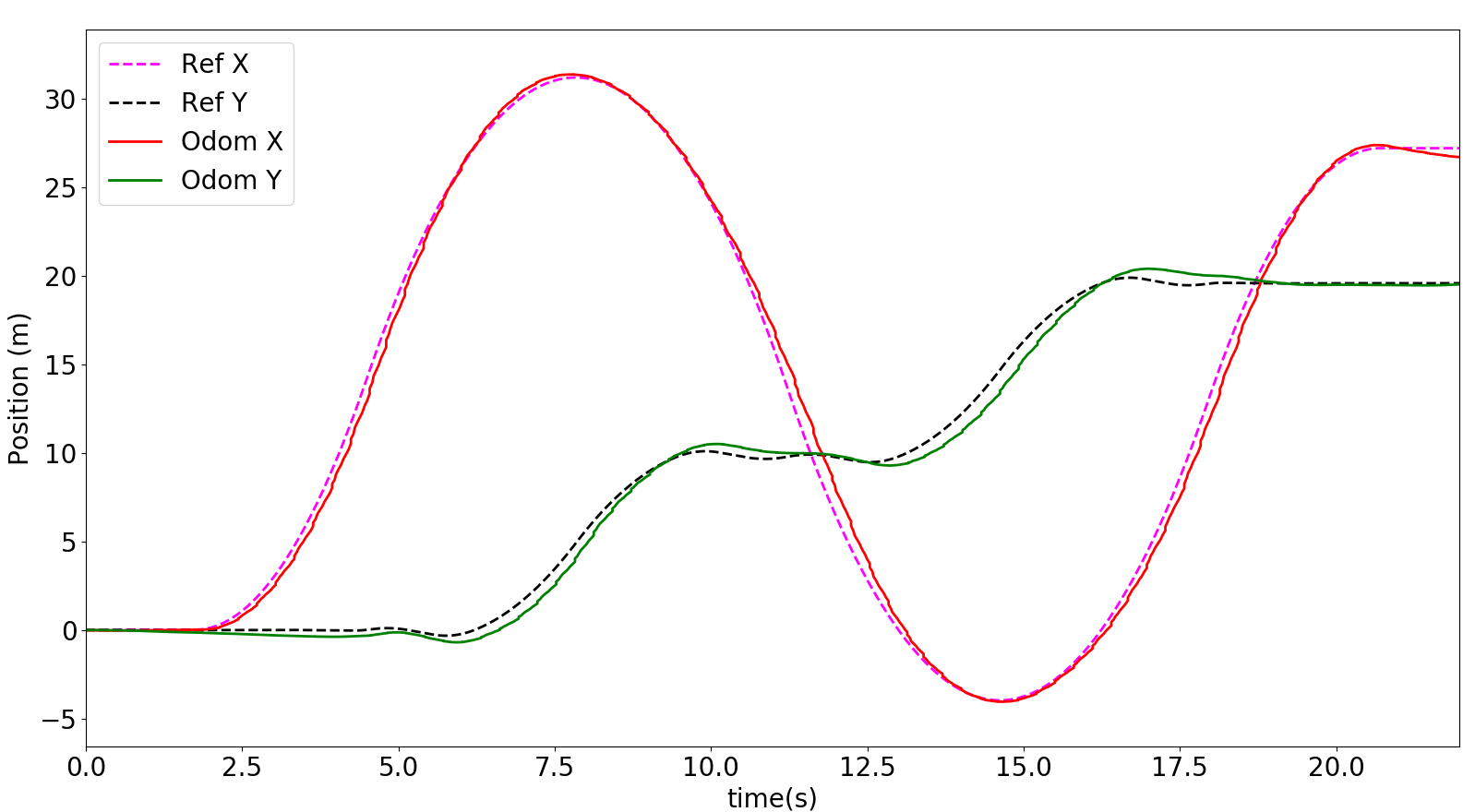}
    \caption{Position tracking performance of the drone }
    \label{fig:pos_tracking}
\end{figure}

\begin{figure}[h]
    \includegraphics[width=3in]{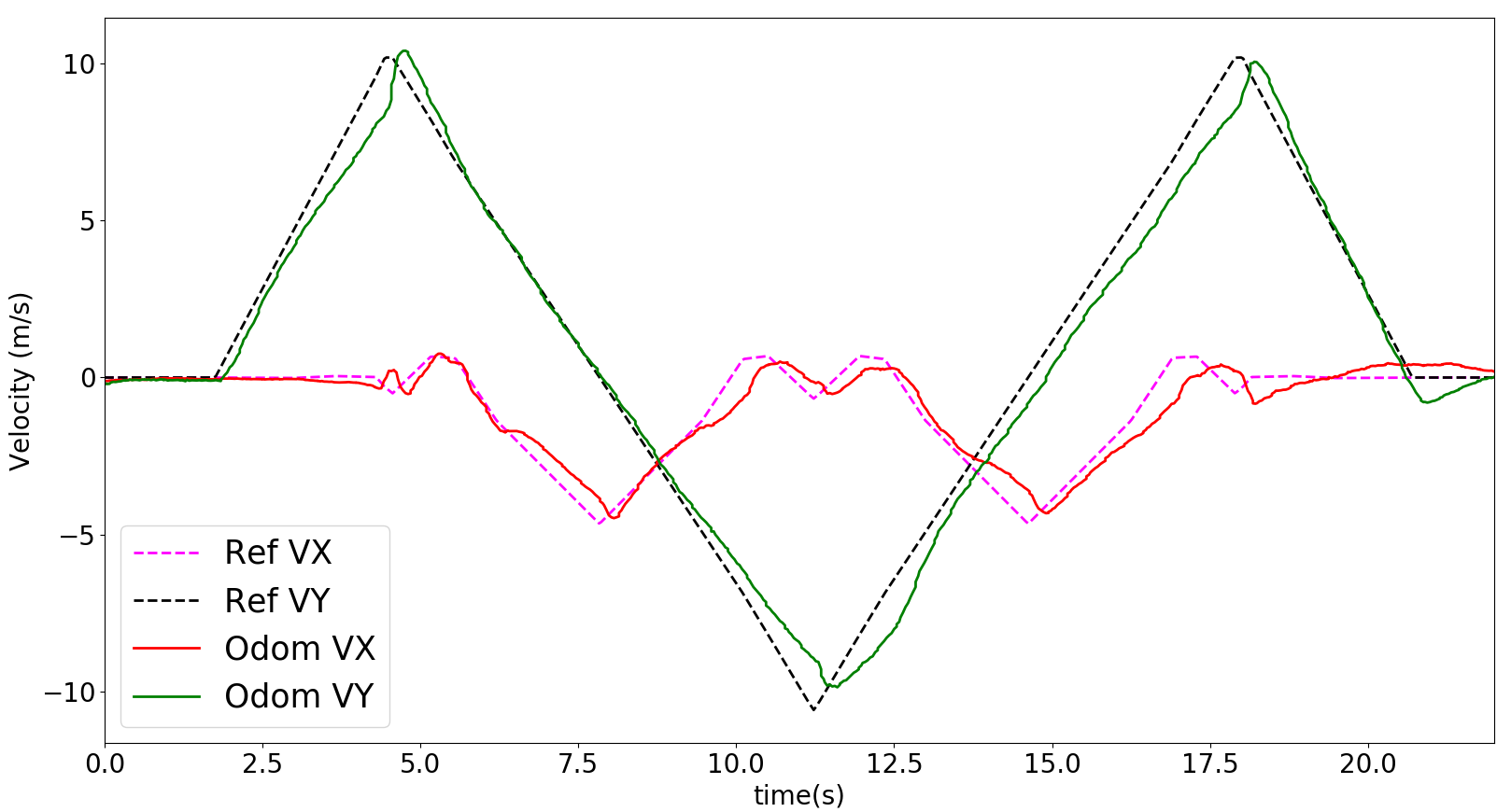}
    \caption{Velocity tracking performance of the drone }
    \label{fig:vel_tracking}
\end{figure}

\begin{figure}[!h]
    \centering
    \includegraphics[width=2.5in]{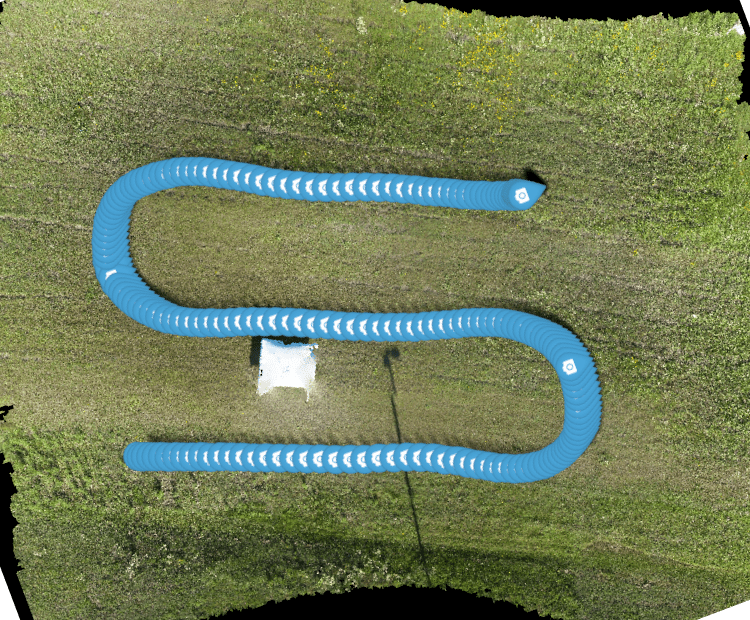}
    \caption{Flight path of surveying UAV overlaid on orthomosaic of survey area}
    \label{fig:campath}
\end{figure}

\begin{figure}[!h]
    \centering
    \includegraphics[width=3in]{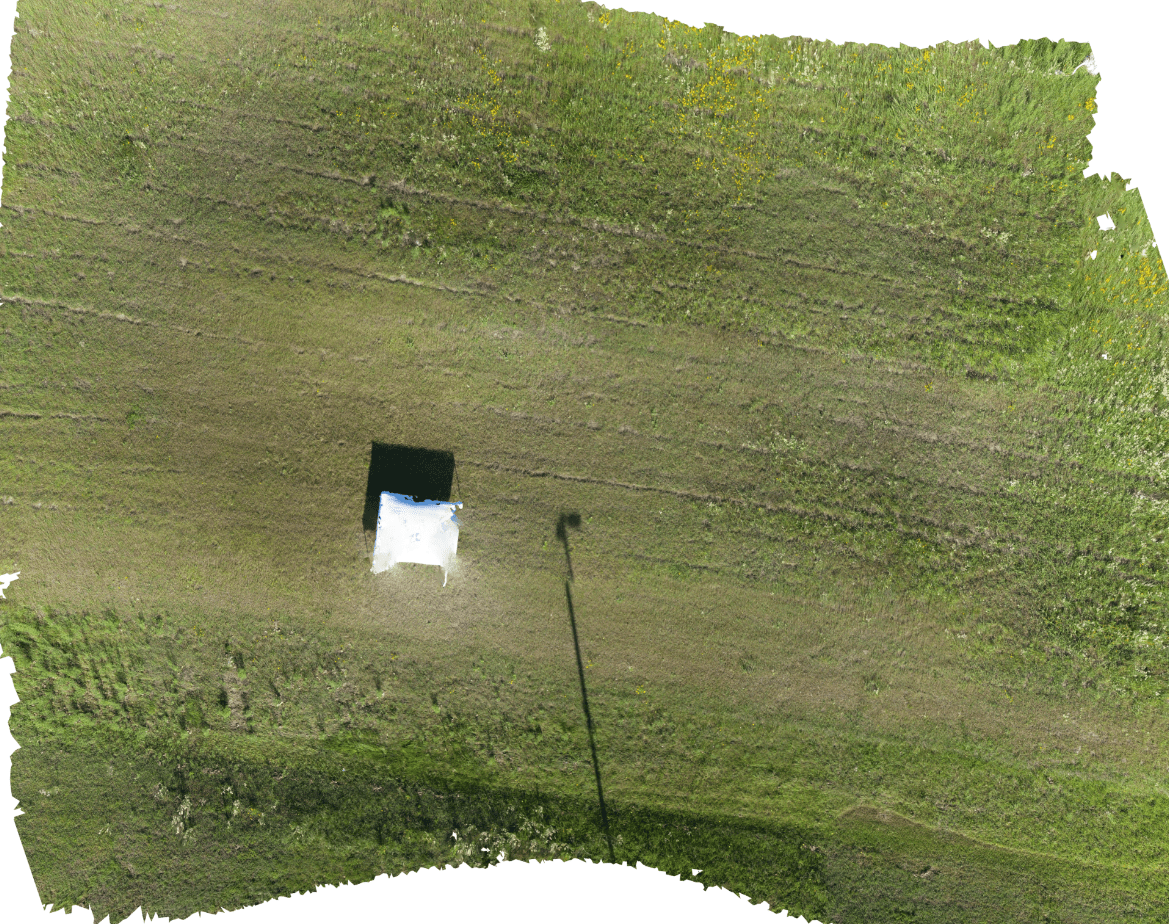}
    \caption{Orthomosaic of the surveyed area}
    \label{fig:orthomosaic}
\end{figure}

Figures \ref{fig:pos_tracking} and \ref{fig:vel_tracking} show the reference trajectory generated for a thrust bound of $10.5N$ along with the trajectory followed by the quadrotor. The quadrotor is able to track the generated trajectory well and the captured images are stitched to generate the orthomosaic shown in fig \ref{fig:orthomosaic} and the corresponding trajectory is shown in figure \ref{fig:campath}.

\section{Conclusion and future work}
\label{sec:conclusions}

In this work, we introduced a time-optimal surveying algorithm which generates a continuous trajectory that uses maximum effort and minimizes aggregate flight time. 
First, we formulate the time optimal survey problem as an NLP by introducing switching points between desired waypoints in the trajectory and discretizing the input to satisfy Pontryagin Minimum Principle. Then, we reduce the computation time for solving the NLP by using an SOCP approximation and using its solution as the initial guess for the NLP optimization. Using the solution to the NLP, we generate position, velocity and acceleration references using direct interpolation and compare its performance to other minimum-time methods in the literature like bang-bang control.

To verify the feasibility of the algorithm in a real world setting, we used a custom built drone to execute a time optimal coverage survey of a known area. The experimental results show that with a well tuned trajectory tracking controller and known camera parameters, one can complete photogrammetry surveys with UAVs in minimal time. The generated references ensured that the drone used maximum actuator input while constraining motion blur in the captured data.

While we have focused on aggregate flight time, total computation time is another aspect that must be optimized for further reduction in time. While using the SOCP approximation greatly reduces the computation time it may not scale with the number of waypoints. The computation time needs to be further accelerated to get as close to real time as possible. This can be addressed by using a data driven approach that learns from the planner described in this work. Investigating this scenario is a direction for possible future work.

\printbibliography
\end{document}